\documentclass[11pt,a4paper]{article}
\usepackage[hyperref]{acl2020}
\usepackage{times}
\usepackage{graphicx}
\usepackage{multirow}
\usepackage{latexsym}
\usepackage{booktabs}
\usepackage{amssymb}
\usepackage{amsmath}
\usepackage{xcolor}
\usepackage{interval}
\usepackage{bbm}
\usepackage{soul}
\usepackage{tabularx}

\definecolor{mypink}{RGB}{252, 199, 199}
\definecolor{myorange}{RGB}{255, 231, 173}
\definecolor{mypurple}{RGB}{244, 224, 254}
\definecolor{mygreen}{RGB}{216, 232, 183}
\newcommand{\rouge}{{\scshape Rouge-1}}
\newcommand{\bartCNN}{$\texttt{BART}_{cnn}$}
\newcommand{\bartXSum}{$\texttt{BART}_{XSum}$}
\newcommand{\ourdata}{{\scshape NeuWS}}

\newcommand{\all}{$\textsc{Bias}$}
\newcommand{\neutral}{\textsc{Neutral}}
\newcommand{\finetune}{$\texttt{BART}_{ft}$}
\newcommand{\tfive}{$\texttt{T5}_{ft}$}
\newcommand{\prop}{$\texttt{BART}_{ft+\mathcal{L}_\texttt{Prop}}$}
\newcommand{\fakenews}{$\texttt{BART}_{ft+\mathcal{L}_\texttt{FakeNews}}$}

\usepackage{microtype}

\aclfinalcopy % Uncomment this line for the final submission
\title{Mitigating Media Bias through Neutral Article Generation}

\author{
Nayeon Lee \quad Yejin Bang \quad Andrea Madotto \quad Pascale Fung \\
Center for Artificial Intelligence Research (CAiRE) \\
Hong Kong University of Science and Technology \\
\texttt {nyleeaa@connect.ust.hk} \\
}

\begin{document}
\maketitle
\begin{abstract}
Media bias can lead to increased political polarization, and thus, the need for automatic mitigation methods is growing. 
Existing mitigation work displays articles from multiple news outlets to provide diverse news coverage, but without neutralizing the bias inherent in each of the displayed articles. 
Therefore, we propose a new task, a \textit{single} neutralized article generation out of \textit{multiple} biased articles, to facilitate more efficient access to balanced and unbiased information.
In this paper, we compile a new dataset (\ourdata), define an automatic evaluation metric, and provide baselines and multiple analyses to serve as a solid starting point for the proposed task. Lastly, we obtain a human evaluation to demonstrate the alignment between our metric and the human judgment. 
% Our dataset, baselines, and code-base will be released to promote more research in this direction.
\end{abstract}

\section{Introduction}
Media bias refers to the bias produced when journalists report about an event in a prejudiced manner or with a slanted viewpoint \cite{gentzkow2006media}. Since news media play a crucial role in shaping public opinion toward various important issues
\cite{de2004effects, mccombs2009news, perse2016media}, bias in media could reinforce the problem of political polarization. 
Due to its potential societal harm, this issue has been extensively studied in social sciences, and there have been both journalistic and computational efforts to detect and analyze media bias \cite{entman1993framing,groseclose2005measure,recasens2013linguistic}. However, computational methods for media bias \textit{mitigation} are still under-explored.

% \begin{table}[]
% \resizebox{\linewidth}{!}{
% \begin{tabular}{l}
% \toprule
% \begin{tabular}[c]{@{}l@{}}\textbf{LEFT:} The Justice Department \colorbox{pink}{rejected a last-ditch} appeal\\ from defense lawyers for the former F.B.I. deputy director \\ Andrew G. McCabe, people familiar with the matter said \\ on Thursday, as prosecutors have appeared close to charging \\ him in a criminal case. [...]\end{tabular} \\ \midrule
% \begin{tabular}[c]{@{}l@{}}\textbf{RIGHT:} The Justice Department on Thursday \colorbox{pink}{notified} \\ lawyers for former FBI Deputy Director Andrew McCabe, \\ a frequent target of President Donald Trump's wrath, that \\ they have rejected his appeal to avoid being criminally \\ charged, according to multiple news outlets \colorbox{pink}{citing multiple} \\ \colorbox{pink}{unnamed sources}. [...]\end{tabular} \\ \midrule
% \begin{tabular}[c]{@{}l@{}}\textbf{CENTER:} Federal prosecutors recommended seeking \\ criminal charges against Andrew McCabe, the former \\ deputy director of the FBI and a frequent target of criticism \\ by President Donald Trump, according to people familiar \\ with the decision Thursday. [...]\end{tabular} \\ \bottomrule
% \end{tabular}%
% }
% \caption{Example}
% \label{table:intro_example}
% \end{table}
\begin{table}[]
\centering
\resizebox{\linewidth}{!}{
\begin{tabular}{l}
\toprule
\begin{tabular}[c]{@{}l@{}}\textbf{\texttt{[Left]}} Dan Bishop’s \colorbox{myorange}{narrow win} suggests that \\ \colorbox{mypurple}{Republicans need to dial up their efforts to keep} \\ \colorbox{mypurple}{the suburbs in 2020.}\end{tabular} \\ \midrule
\begin{tabular}[c]{@{}l@{}}\textbf{\texttt{[Right]}} Republicans scored an \colorbox{myorange}{important victory} \\ in  North Carolina Tuesday as \colorbox{mypurple}{President Trump  }\\ \colorbox{mypurple}{helped them} hang on to a GOP congressional \\ seat in a \colorbox{myorange}{closely-watched} special election.\end{tabular} \\ 
\midrule
\begin{tabular}[c]{@{}l@{}}\textbf{\texttt{[Neutral]}} Republican Dan Bishop won an \\ special election in North Carolina on Tuesday.\end{tabular} \\ 
\bottomrule
\end{tabular}%
}
\caption{Illustration of news articles from different political ideologies (Left vs. Right). The highlighted spans refer to \colorbox{myorange}{lexical} and \colorbox{mypurple}{informational} bias.}
\label{table:intro_example}
\end{table}

Currently, the main computational approach to mitigating media bias is aggregation of multiple news articles to provide a comprehensive reporting with the additional analysis of news outlets
\cite{park2009newscube,sides2018media,zhang2019tanbih}. In this way, readers can access diverse news coverage, however, at the cost of reading more articles than what they would normally read.
% this method requires consumers to read multiple articles that can be time-consuming.
Moreover, since bias within each individual article is not neutralized, it could still undesirably sway readers' views.
According to \citeauthor{bail2018exposure}, exposure to opposing political views can actually reinforce political polarization. Therefore, presenting articles from different stances alone cannot entirely solve the problem.

As a remedy, we take a step forward from the news aggregation approach and propose a new task -- generating a \textbf{single} neutralized article out of \textbf{multiple} biased articles on the same event (\textit{multi-document neutralization}\footnote{For ease of writing, we refer to our task as ``neutralization'' in the rest of the paper}). 
% The reportings from conflicting news outlets still share the ``same set of underlying facts'', however, differ by deliberate omission of certain fact and biased/subjective choice of words to frame same event in direction of preferred stance~\cite{gentzkow2006media}. 
The articles from conflicting news outlets still share the ``same set of underlying facts'', however, convey different impression of same event through deliberate omission of certain fact and slanted choice of words~\cite{gentzkow2006media}. 
With an automatic method to extract and aggregate neutral information from multiple articles, the public could easily access more unbiased news information, leading to reduced risk of political polarization. 

In this work, we formulate our new task by i) constructing a weakly-labeled news neutralization dataset, which we call \ourdata, ii) defining a general model setup, and iii) designing an automatic metric called the \neutral~score to evaluate the success of the neutralization. Then, we establish a solid baselines by leveraging large pre-trained model for \textit{neutralization}. Lastly, we conduct human evaluation to examine how aligned our \neutral~score is with human judgement, and provide interesting insights from additional experimental analysis that suggests potential directions for future work.

\section{Related Work}
\paragraph{Media Bias Detection and Prediction} 
Media bias has been studied extensively in various fields such as social science, economics and political science, and various measures have been used to analyze the political preference of news outlets~\cite{groseclose2005measure,miller2001spiral, park2011contrasting,gentzkow2010drives,haselmayer2017sentiment}. 
For instance, \citeauthor{gentzkow2010drives} count the frequency of slanted words within articles. 
In natural language processing (NLP), computational approaches for detecting media bias consider lexical bias, which is linguistic cues that induce bias in political text~\cite{recasens2013linguistic,yano2010shedding,lee2019team, hamborg2019illegal}. 
While these methods specifically focus on the lexical aspect of media bias, our work attempts to address media bias more comprehensively.

% informational bias focus 
As highlighted by \citeauthor{fan2019plain}, media bias also has an informational aspect due to framing bias, which is selective reporting of an event to sway readers' opinions -- e.g., omission of crucial facts or choice of words ~\cite{entman1993framing, entman2007framing, gentzkow2006media}. 
Efforts related to informational bias~\cite{park2011contrasting,fan2019plain} are constrained to detection tasks. In this work, we attempt to tackle mitigation of media bias (both lexical and informational bias).

% Beyond the bias-inducing words, framing is often selected by media to sway readers opinion. \citeauthor{entman1993framing} defined framing as selectively present and emphasize a piece of reality of an event in text in a way of leading to desired interpretation on the event. In news articles, it does not take a stance explicitly~\cite{miller2001spiral}, but rather selective omission of certain fact, choice of words, and varying credibility ascribed to the primary source are used to frame an event in direction of preferred stance~\cite{entman2007framing, gentzkow2006media}. 
% In recent years, computational approaches on framing have been made. 
% \citeauthor{card2015media} clearly categorized frames and released frame corpus. 
% Automated identifications of bias based on framing theory has been made by utilizing key opponents of an issue and contrasting disputants among articles~\cite{park2011contrasting} and focusing on word choice and labeling~\cite{hamborg2019illegal}.

\paragraph{Media Bias Mitigation}
News aggregation, displaying articles from different news outlets on a particular topic (e.g., Google News,\footnote{https://news.google.com/} Yahoo News\footnote{https://news.yahoo.com/}), is the most common approach in NLP to mitigate media bias, but it still has limitations~\cite{hamborg2019automated}.
Thus, multiple approaches have proposed to provide additional information~\cite{laban2017newslens}, such as automatically classified multiple view points~\cite{park2009newscube}, multinational perspectives~\cite{hamborg2017matrix}, and detailed media profiles~\cite{zhang2019tanbih}. \textit{Allsides.com}\footnote{https://www.allsides.com/} provides bias ratings of each news outlet alongside balanced political coverage. However, they focus on making news consumers more aware of what they are reading. Thus far, there has been no attempt to automatically aggregate biased articles to produce a single neutralized article.

\paragraph{Controlled Text Generation}
One line of work under controlled text generation tries to de-bias or neutralize text~\cite{dathathri2019plug, pryzant2020automatically, ma2020powertransformer}. Specifically, efforts were made to reduce toxicity~\cite{dathathri2019plug}, implicit social bias~\cite{ma2020powertransformer} or subjectivity~\cite{pryzant2020automatically} in generations.
Our work differs from theirs in two ways: 1) Neutralization in the previous works is close to revision of single input text with a focus on style, while our work is neutralized aggregation of multiple texts. 2) We focus on media bias. To the best of our knowledge, we are the first to attempt politically neutralized news article generation.
% \citeauthor{dathathri2019plug} try to reduce the toxicity of a generation, \cite{pryzant2020automatically} aim to reduce subjectivity for general application purpose, and \citeauthor{ma2020powertransformer} attempts to stylistically rewrite a given text to correct the implicit biases in character portrayals.
% \cite{shwartz2020neural} - explores the reporting bias that exists in large pretrained language models 
% requires neutralized piece of reporting out of multiple sources while delivering essential and unbiased information regarding an issue.

\paragraph{Hallucination} 
Recent studies have shown that neural sequence models can suffer from hallucination of additional content, not supported by the input, as result, adding factual inaccuracy to the generation of abstractive summarization models. To address this problem, many researchers proposed methods to measure factual inconsistency~\cite{holtzman2019curious, kryscinski2019evaluating, zhou2020detecting, lux2020truth, gabriel2020go}, and to correct them \cite{zhao2020reducing, cao2020factual, dong2020multi}. While these works focus on the factual inaccuracy and inconsistency, we focus on the bias that is not factually incorrect but can still pose a problem due to the way how it affects the readers' opinion.
% The most ideal output of neural generation would be to be free from both hallucination and bias, but the focus of our work is on the latter part only.

\begin{table*}[]
\resizebox{\linewidth}{!}{
    \begin{tabular}{cl}
    \toprule
    %  & \multicolumn{1}{c}{\textbf{Text}} \\ 
    %  \midrule
    Event & Democratic presidential candidates ask to see full Mueller report
    \\\midrule\midrule
    Left & \begin{tabular}[c]{@{}l@{}}Democrats want access to special counsel Robert Mueller’s investigation into Russian interference in \\ the 2016 presidential election \colorbox{mypurple}{before President Donald Trump has a chance to interfere.} {[}...{]} Sen. Mark \\ Warner said in a statement: \colorbox{mypurple}{“Any attempt by the Trump Administration to cover up the results of this} \\ \colorbox{mypurple}{investigation into Russia’s attack on our democracy would be unacceptable.”}\end{tabular} \\ \midrule
    Right & \begin{tabular}[c]{@{}l@{}}Democratic presidential candidates \colorbox{myorange}{wasted no time} Friday evening demanding the immediate public\\ release of the long-awaited report from Robert S. Mueller III. {[}...{]} Several candidates, in calling for\\ the swift release of the report, also \colorbox{mypurple}{sought to gather new supporters and their email addresses} by \\ putting out “\colorbox{myorange}{petitions}” calling for complete transparency from the Justice Department.\end{tabular} \\ 
    % \midrule
    % Neutral & \begin{tabular}[c]{@{}l@{}}A fight brewed between Democrats and Republicans over the public release of Special Counsel Robert \\ Mueller’s report on Russian meddling in the 2016 U.S. election, while President Donald Trump kept \\ up attacks on his critics on Monday.\\ As the Republican chairman called for an investigation into the origins of the probe of any Trump \\ campaign links with Russians, the Senate leader blocked a second attempt by Democrats to pass \\ a measure aimed at pushing the Justice Department into full disclosure of the report. \end{tabular} \\ 
    \bottomrule
    \end{tabular}%
}
\caption{Illustration of \colorbox{mypurple}{informational bias spans} and \colorbox{myorange}{lexical bias spans} from BASIL dataset~\cite{fan2019plain}}
\label{table:basil_example}
\end{table*}

\section{Task Formulation}
\subsection{Task Definition}
\label{sec:task_definition}
The main objective of this work is to \textit {neutralize} biased news articles from two different news outlets (left-winged and right-winged\footnote{For simplicity, we use two biased articles, but this can be extended to more than two articles.}) into a single neutral article which (i) retains as much information as possible and (ii) eliminates as much bias as possible from the input articles.

\paragraph{Media Bias Definition} We follow the categorization and definition of media bias from \citeauthor{fan2019plain} There are two types of media bias: \textit{lexical bias}, which refers to the writing style or linguistic attributes that may mislead readers, and \textit{informational bias}, which refers to tangential or speculative information pieces to sway the minds of readers. Such biases can make an article convey a different impression of what actually happened~\cite{gentzkow2006media}. Ideally, a neutral article should avoid both types of bias by using a neutral tone and including balanced information without preference towards any particular stance or target.

\begin{table}[]
\centering
\small
\begin{tabular}{lccc|c}
% \toprule
% \multicolumn{1}{c}{Ideology} & \multicolumn{1}{c}{Train/Val} & Test \\
% \midrule
% % \cmidrule(lr){1-1}\cmidrule(lr){2-3}\cmidrule(lr){4-4}
% % \multicolumn{1}{c}{Type} & Full Article & Snippet & Full Article \\ \midrule\midrule
% Left & 580 & 100 \\
% Right & 580 & 100 \\
% Center & 580 & 100 \\ \midrule
% Total & 1,740 & 300 \\ \bottomrule
\toprule
 & Left & Right & Center & Total \\ \midrule
Train/Valid & 580 & 580 & 580 & 1,740 \\
Test & 100 & 100 & 100 & 300 \\ \bottomrule
\end{tabular}
\caption{\ourdata~data statistics. Article triplet refers to the set of \{left, right, center\} news for each events.}
\label{table:stat}
\end{table}

\subsection{\underline{\textsc{Neu}}tralize Ne\underline{\textsc{WS}} Dataset (\ourdata)}
The neutralization task requires a dataset that consists of politically opposing source articles and neutral target article, reporting about same event. 
We, therefore, build a weakly-labeled data from article triplets consisting of articles from politically left, right and center publishers. The dataset language is English, and mainly focuses on the U.S. political events only.

We term our dataset to be weakly-labeled on the following basis. First, there is no single answer to writing a neutral news, thus, there cannot be ``the gold'' neutral article to optimize for. Second, since all news requires editorial judgements on what is the ``important'' information to report, it is possible that even news articles from the most neutral publisher contain some bias. Note that, the literature still considers center publishers to be bias-free, especially in comparison to other hyper-partisan publishers~\cite{baum2008new}. 

For the political orientations of the publishers, we rely on the Media Bias Ratings\footnote{https://www.allsides.com/media-bias/media-bias-ratings}, which uses editorial reviews, blind bias surveys (10,000+ community participants), independent reviews, and third party research to correctly judge the political stance of various publishers. 

\paragraph{Train/Valid Set} 
To construct this train/valid set, we first crawled the URLs of article triplets from Allsides.com, which displays news coverage of events from left, right, and center publishers. Then, we built custom news crawlers to obtain the full article content from the collected URLs. In total, we collected 1,740 full articles which were compiled into 580 article triplets. The data statistics are listed in Table~\ref{table:stat}.

\paragraph{Test Set} For test set, we utilize a subset of BASIL dataset \cite{fan2019plain} which contains sentence-level annotation of media bias within news articles. These annotation of bias spans are key to the measurement of the neutralization performance. This dataset is only used in testset because it is small (100 samples) to be split into train/val/test.

We extend the BASIL dataset by adding center articles.
BASIL dataset originally consists of article triplets from Huffington Post (left-wing), Fox News (right-wing) and New York Times (left-wing). 
The New York Times is the outlet closest to center among the three. However, its political leaning is still considered pro-Liberal~\cite{puglisi2011being, chiang2011media}. Therefore, we \textit{replace} the New York Times articles with those from center publishers (e.g. Reuters, BBC) that report on the same event. To ensure that the newly collected center news articles are covering the same event as the left/right articles, we manually confirmed the content and the publication dates.

\paragraph{Notations} Here, we introduce the notations used throughout the paper. We denote the two biased articles with $X^{(l)}=\{x^{(l)}_1, \cdots, x^{(l)}_L\}$ and $X^{(r)}=\{x^{(r)}_1, \cdots, x^{(r)}_R\}$, and a center article as $X^{(c)}=\{x^{(c)}_1, \cdots, x^{(c)}_C\}$, where $x$ represents a token. These three articles form one triplet of our dataset $D=\{(X_i^{(l)},X_i^{(r)},X_i^{(c)})\}_{i=1}^{N}$.
Then, we denote the set of annotated bias span $B^{(l)}=\{ X^{(l)}_{i,j}|1\leq i\leq j \leq z\}$ and $B^{(r)}=\{X^{(r)}_{i,j}|1\leq i\leq j \leq m\}$ as the bias sub-strings in $X^{(l)}$ and $X^{(r)}$ respectively. An example of bias spans is shown in Table~\ref{table:basil_example}.

\subsection{General Model Architecture}
% \paragraph{Input} The input to the model is pre-processed by concatenating two biased articles with the special ``[SEP]'' token in between. For half of the time, left articles comes first, and for the other half of the time, right articles comes first. This was done to ensure that the model does not learnt unintended patterns based on particular political ideology.
\paragraph{Model} 
Following the current state-of-the-art in sequence-to-sequence modelling, we propose to firstly encode the concatenation of the two articles and to then generate a neutralized article token-by-token using a decoder.
Hence, given the two articles $X^{(l)}$ and $X^{(r)}$ as a single sequence of tokens, the encoder processes the input as follows,:
\begin{equation}
    H = \mathrm{ENC}_{\theta}(X^{(l)};X^{(r)}),
\end{equation}
where $H\in\mathbb{R}^{z+n\times d}$ and $d$ is the hidden feature size. Note that the conversion between tokens and embedding is done directly in the encoder $(ENC)$. This hidden representation is then passed to the decoder, which generates an article $\hat{X}^{(n)}$ token-by-token in an autoregressive manner. More formally,
\begin{equation}
    \hat{X}^{(n)} = \mathrm{DEC}_{\theta}(H),
\end{equation}

In the experiment section (Section~\ref{sec:experiment}), we provide more details on how we train this general model.

\begin{table*}[t]
    \begin{minipage}{.6\linewidth}
        \begingroup
            \resizebox{1.0\textwidth}{!}{
            \begin{tabular}{llccc}
            \toprule
            & Baseline & \rouge & \all & \textbf{\neutral} \\ \midrule \midrule
            \multirow{2}{*}{A. Zero-shot} 
            & \bartCNN & 16.93 & 30\% & 11.85 \\
            & \bartXSum & 7.79 & 6\% & 7.32 \\ 
            \midrule
            \multirow{2}{*}{B. Fine-tuning } 
            & \finetune~ & 38.11 & 42\% & 22.10 \\
            & \tfive~ & 18.72 & 38\% & 11.61 \\ 
            \midrule
            \multirow{2}{*}{C. \finetune + Disinfo} 
            & \fakenews & 47.90 & 54\% & 22.03 \\
            & \prop & 46.86 & 23\% & \textbf{36.08} \\
            \bottomrule
            \end{tabular}
            }
            % \vspace{24pt}
            \caption{Baseline experiment results using beam-4 decoding technique. For \rouge~ Recall and \neutral, \textit{higher} number is better. For \all, \textit{lower} number is better.} 
            \label{table:baseline_results}
        \endgroup
    \end{minipage}%
    \hspace{8pt}
    \begin{minipage}{.38\linewidth}
        \centering
        \begingroup
        \includegraphics[width=\linewidth]{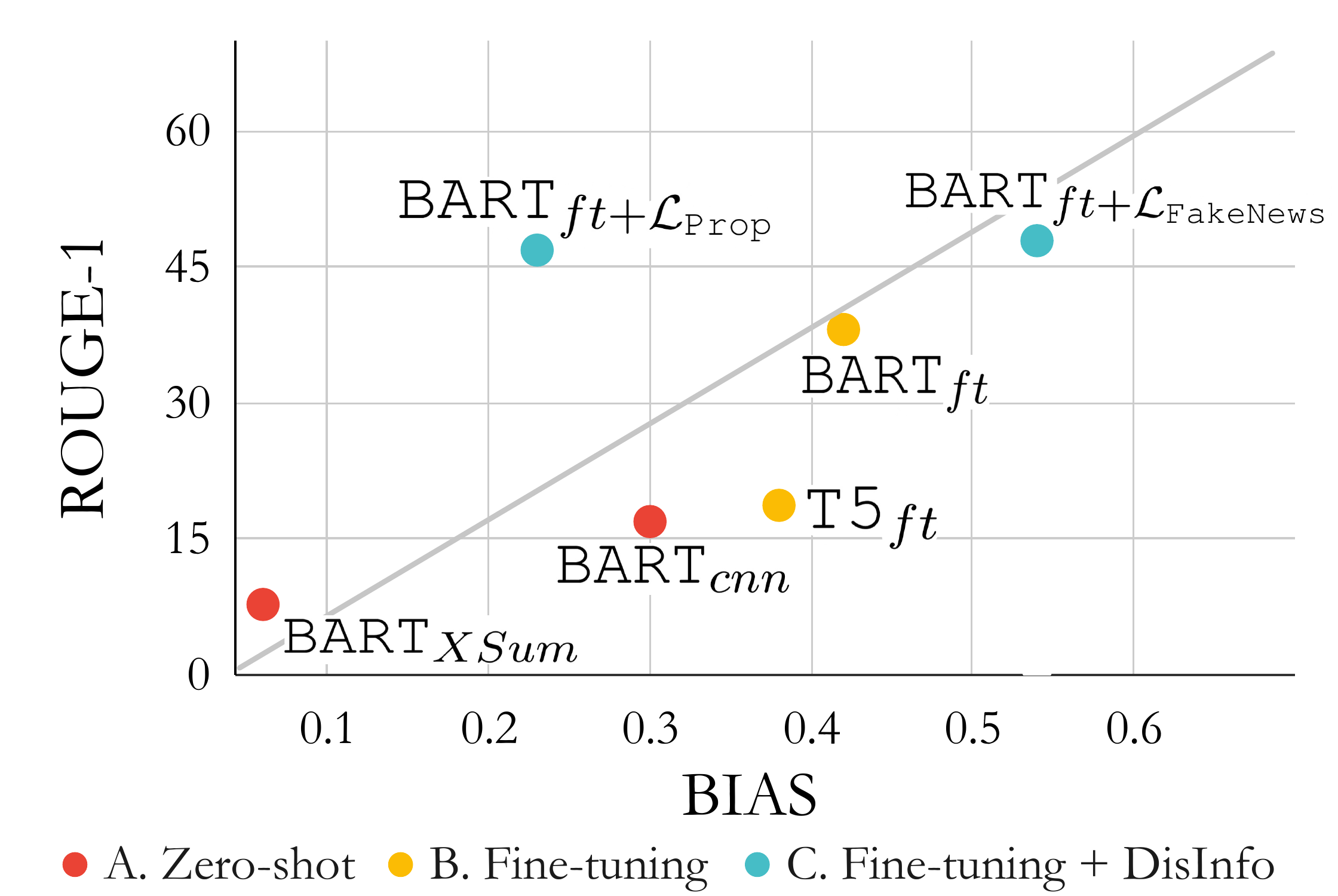}
        \vspace{-20pt}
        \captionof{figure}{Plot of experiment results. The closer to the top-left, the better.}
        \label{fig:result_illustration}
        \endgroup
    \end{minipage}%
\end{table*}
% \begin{table*}[]
% \centering
% \begin{tabular}{llccc}
% \toprule
%  & Baseline & \rouge & \all & \textbf{\neutral} \\ \midrule \midrule
% \multirow{2}{*}{A. Zero-shot} 
% & \bartCNN & 16.93 & 30\% & 11.85 \\
% & \bartXSum & 7.79 & 6\% & 7.32 \\ 
% \midrule
% \multirow{2}{*}{B. Fine-tuning Seq2Seq} 
%  & \finetune~ & 38.11 & 42\% & 22.10 \\
%  & \tfive~ & 18.72 & 38\% & 11.61 \\ 
%  \midrule
% \multirow{2}{*}{C. \finetune + Disinfo Loss} 
% & \fakenews & 47.90 & 54\% & 22.03 \\
% & \prop & 46.86 & 23\% & \textbf{36.08} \\
%  \bottomrule
% \end{tabular}
% \caption{Baseline experiment results using beam-4 decoding technique. For \rouge~ Recall and \neutral, \textit{higher} number is better. For \all, \textit{lower} number is better. }
% \label{table:baseline_results}
% \end{table*}

\subsection{Neutralization metric}
\label{section:metric}
Two important goals for successful neutralization are to minimize bias and to maximize information recall. Therefore, we introduce ways to measure each aspect and combine the both to serve as the final neutralization score (\neutral). 
% The implementation of these metrics will be released to facilitate easy extension of our work by others.

\paragraph{Bias Metric} 
The most important criterion for success is to assess whether the generated article $\hat{X}^{(n)}$ manages to filter out bias spans ($B^{(l)}$ and $B^{(r)}$) originally existing in the input articles.

To quantify this, we obtain generated neutralized-articles for the whole test set ($\{\hat{X}_i^{(n)}\}_{i=1}^{N}$), and measure the ratio that still contains at least one bias span. The lower the ratio, the better the neutralization performance (i.e., less bias spans). 
% By leveraging the gold annotation of the bias spans, we quantify how successful a given model is in avoiding bias spans when generating the neutralized reporting. 
Formally, we define the bias score as follows:
\begin{align}
    \textsc{Bias}  = \frac{1}{N}\sum_{k=1}^{N} \textsc{BiasExists}(B_k^{(A)},\hat{X}_k^{(n)}), 
\end{align}
where $B^{(A)}= \{B^{(l)} \cup B^{(r)}\}$ refers to the union of the bias spans from left-wing and right-wing articles, and \textsc{BiasExists} refers to a function that identifies the existence of bias in a given generated article ($\hat{X}$):
\begin{equation}
    \textsc{BiasExists}(B,\hat{X}) = \\
    \begin{cases}
      1 & \exists X_{i,j}\in B \ \mathrm{and} \\ & X_{i,j} \in \hat{X} \\
      0 & \text{Otherwise.}
    \end{cases}     
\end{equation}

% The implementation of $X_{i,j} \in \hat{X}$ can be done in many different ways.We explored both semantic-match and  lexical-match, and concluded that lexical-match is simpler, faster and more error-free. Thus, we use the lexical-match for our experiments, but both versions of implementation will be released.

\paragraph{Information Recall} One of the easiest way of deceiving the \all~metric is to generate random text that has nothing to do with the biased source articles. Therefore, it is crucial to also ensure that key information is being retained while removing the bias.

We adopt the \rouge~\cite{lin2004rouge} recall score 
% to approximate the coverage of key facts in the neutralized generation. 
between generated neutralized-article, $\hat{X}^{(n)}$, and center article, $X^{(c)}$ to measure the information coverage. The higher the \rouge~score, the better the information covered by the neutralized generation. We mainly report unigram-based \rouge~for simplicity,
% \yeon{(write better) as it was more sensitive/ more discriminating/ more informative due to its forgiving nature}
but {\scshape{Rouge-2}} and {\scshape{Rouge-L}} are also reported in the appendix for interested readers.

\paragraph{Neutralization Score (\neutral)}
Since the ultimate goal is to optimize for both the information recall score and bias score, we define a single neutralization score by combining the two: 
\begin{equation}
    \textsc{Neutral} = \textsc{\rouge}~\times (1 - \textsc{Bias}).
\end{equation}
Through multiplication, both scores get equal weighting in the final metric. 
% We do not only focus on the removal of bias, but also the maximal delivery of the important information.

\section{Models and Experiments}
\label{sec:experiment}
Since neutralized article generation is a new task with no baseline, we propose three different baseline methods to serve as reference points.

\subsection{Zero-shot Using Summarization Model}
Summarization, the task of producing a shorter version of one or several documents that preserves most of the input's meaning, has a similar setup to our proposed neutralization task. 
Therefore, we investigate the zero-shot neutralization performance of two strong BART-based~\cite{lewis2019bart} summarization models by utilizing their pre-trained weights to initialize our encoder ($\mathrm{ENC}_{\theta}$) and decoder ($\mathrm{DEC}_{\theta}$). We test with two version of BART trained on the CNN/DailyMail~\cite{hermann2015teaching} and XSum~\cite{narayan2018don} dataset -- \bartCNN~and \bartXSum.

\subsection{Fine-tuning Seq2Seq Model} 
\label{sec:seq2seq}
Many Transformer-based~\cite{vaswani2017attention} pre-trained language models~\cite{raffel2019exploring,lewis2019bart} achieved excellent performance in downstream tasks through simple fine-tuning with, small, task-specific data. 
Therefore, we fine-tune the encoder-decoder parameters of pre-trained Seq2seq models by minimizing the negative log-likelihood over the training set $D$. We experiment with the following two pre-trained models:
\begin{itemize}
    \item \finetune: pre-trained BART-large model fine-tuned with \ourdata~data.
    \item \tfive: pre-trained T5-base~\cite{raffel2019exploring} model fine-tuned with \ourdata~data.
\end{itemize}

\subsection{Incorporating Disinformation Loss}
Next, we explore adding an additional loss when fine-tuning the Seq2Seq model. 
% The extra disinformation classification loss allows a model to learn about hyper-partisan writing, so the model is prevented from generating such text.
% The extra disinformation classification loss will encourage the model to learn about hyper-partisan writing, which we believe could help the model to restrain from generating such text.
We encourage the model to learn about hyper-partisanship writing and, as result, learn to avoid generating text alike.
This is done by incorporating an additional classification head on top of the encoder ($\mathrm{ENC}_{\theta}$) to jointly optimize for this classification cross-entropy loss with the original negative log-likelihood loss from the decoder ($\mathrm{DEC}_{\theta}$). 
In our experiments, the following settings are explored:
\begin{itemize}
    \item \fakenews: pre-trained BART-large jointly fine-tuned on \ourdata~and fake news dataset \cite{potthast2017stylometric}.
    \item \prop: pre-trained BART-large jointly fine-tuned on \ourdata~and propagandistic sentence detection task \cite{da2019fine}, which is to classify whether a given sentence contains any propagandistic technique (e.g., ``Name calling'', ``Appeal to fear'').
\end{itemize}
% We denote the cross entropy loss function induced by the classification loss as $L_{fakenews}$, and $L_{prop}$

\subsection{Experimental Details}
All our experimental codes are based on the HuggingFace library~\cite{wolf-etal-2020-transformers}. During training, and across models, we used the following hyper-parameters: $10$ epoch size, $3e-5$ learning rate and a batch size of $8$. We did not do hyper-parameters tuning since our objective is to provide various baselines and analysis. Training run-time for all of our experiments are fast ($<6$hr). 
No pre-processing of the text was made, except for the concatenation of left and right articles with special token [SEP] in the middle (i.e. `` left-article [SEP] right-article ''). When concatenating, we ensured the first half to begin with left-articles and the other half to begin with right-articles. This was done to avoid any unintended bias from the ordering of the ideology in the input. 
% All BART based models in our experiments use of pre-trained BART-large model, which has 406M parameters and T5-base has 220M parameters.

\section{Results}
All the experimental results are reported in Table~\ref{table:baseline_results}.

\paragraph{A. Zero-shot} From Table~\ref{table:baseline_results}, we can observe that the zero-shot performance of both summarization models on the proposed neutralization task are poor in terms of \neutral. One possible reason is due to the difference in the training data distribution, especially regarding the characteristics of the target generation. In fact, summarization tasks focus on obtaining \textit{concise} and representative ``summary'' whereas our task focuses on obtaining \textit{neutral} and representative article. Within the summarization models, we can observe that different training datasets (CNN vs XSum) lead to different neutralization performance ($11.85$ vs. $7.32$). 
% Quantitatively speaking, only $1.01 \pm 0.10$ and  $3.06 \pm 0.80$ sentences are generated as neutralized articles from \bartXSum~and \bartCNN~ respectively.

It is important to note that the final \neutral~score of \bartXSum~is very low ($7.32$) despite having the lowest \all~score ($6\%$). The reason behind \bartXSum's low bias score is its extremely short generation length ($1.01 \pm 0.10$ sentence on average). By design, our \neutral~metric includes the \rouge~score as well, which serves as a counteractor in such scenario. This illustrates the effectiveness of considering both the comprehensiveness and neutrality of the generated article to avoid pitfalls.

\paragraph{B. Fine-tuning} 
We compare and report two fine-tuned Seq2Seq models. It is evident that the choice of the base model (i.e., \finetune~vs. \tfive) greatly affects the neutralization performance after the fine-tuning. \finetune~achieves double the \neutral~score compared to the zero-shot baselines, but \tfive~hardly shows any improvement. A likely explanation for this observation would be the difference in the parameter size of these two base models. BART-large has 406M parameters when T5-base only has 220M, so \finetune~has access to more rich features and bigger model capacity. 
% A possible explanation for this observation could be that the pre-trained BART has more useful feature/knowledge that helped achieve better neutralization result through transfer learning. 

From the zero-shot and fine-tuning experiments, we can observe a weak positive correlation between the \rouge~scores and the \all~scores - i.e. lowest \rouge~result ($7.79$) with the lowest \all~score ($6\%$) and highest \rouge~($38.11$) with the highest \all~score ($42\%$).
This is because these baseline models failed to select the important information in a neutral manner from input articles.
With this positively correlating pattern, achieving even the highest performance in one metric would not lead to a good neutralization score. To illustrate, achieving 100.00 \rouge~can still result in 0.00 \neutral~score if \all~is also 100.
Therefore, more sophisticated models are required to correctly identify as much important information as possible while avoiding selecting bias spans at the same time.
% However, improving either \rouge~or \all~ alone cannot lead to the best \neutral, instead, both scores need to concurrently be improved. 

\paragraph{C. Incorporating Disinformation Loss}
In this section, we analyze the impact of adding additional disinformation loss to the \finetune~model's encoder. This experiment is based on \finetune~since it is the best performing model from the earlier fine-tuning experiment. 
The most notable result is that the result of \prop~does not follow the concerning pattern observed earlier.
This is illustrated in Fig~\ref{fig:result_illustration}, where \prop~is leaning toward the left-top.
% Both ``$\mathcal{L}_{\texttt{FakeNews}}$'' and ``$\mathcal{L}_{\texttt{Propaganda}}$'' experiments have \rouge~scores, but very different \all~scores and \neutral scores.
Both the ``\fakenews'' and ``\prop'' experiments have similar \rouge~scores with only 1.04 difference, but with a clear distinction in \all~scores and, as result, different \neutral~scores.
Propaganda loss is empirically shown to be more effective in reducing the media bias in the generation, achieving the highest \neutral~score ($36.08$).

\begin{table}[]
\resizebox{\linewidth}{!}{
    \begin{tabular}{lccccc}
    \toprule
    \multirow{2}{*}{\textbf{\begin{tabular}[c]{@{}l@{}}Decoding \\ Technique\end{tabular}}} & \multirow{2}{*}{\textbf{Hparam}} & \multirow{2}{*}{\textbf{\rouge}} & \multirow{2}{*}{\textbf{\all}} & \multicolumn{2}{c}{\textbf{\neutral}} \\ \cmidrule(lr){5-6}
     &  &  &  & \textbf{Indiv.} & \textbf{Avg.} \\ \midrule \midrule
    \multirow{2}{*}{BEAM} & B:4 & 46.86 & 23\% & 36.08 & \multirow{2}{*}{35.61} \\
     & B:5 & 46.85 & 25\% & 35.14 &  \\ \midrule
    %  & B:6 & 47.42 & 25\% & 35.57 &  \\ \midrule
    \multirow{2}{*}{Top-k} 
     & k:5,t:1.0 & 36.51 & 21\% & 28.84 & \multirow{2}{*}{29.27} \\
    %  & k:5,t:1.0 & 47.18 & 28\% & 33.97 & \multirow{2}{*}{31.90} \\
     & k:50,t:1.0 & 36.67 & 19\% & 29.70 &  \\ \midrule
    %  & k:50,t:1.0 & 46.80 & 25\% & 35.10 &  \\ \midrule
    \multirow{2}{*}{Top-p} 
    %  & p:0.8,t:0.9 & 43.88 & 11\% & 39.05 & \multirow{4}{*}{\textbf{38.26}} \\
     & p:0.8,t:1.0 & 44.93 & 17\% & 37.29 & \multirow{2}{*}{\textbf{36.63}}  \\
    %  & \multicolumn{1}{l}{p:0.9,t:0.9} & 44.29 & \textbf{8\%} & \textbf{40.75} &  \\
     & \multicolumn{1}{l}{p:0.9,t:1.0} & 43.33 & 17\% & 35.96 &  \\
    \bottomrule
    \end{tabular} %
}
\caption{Effect of decoding technique choices on the \neutral. All results are obtained from \prop~.}
\label{table:decoding_analysis}
\end{table}

\begin{table*}[t]
\small
\begin{tabular}{p{15cm}}
\toprule
\textbf{Event 1: Obama willing to use executive orders on guns} \\ \midrule

\textbf{{[}\finetune{]}}: President Obama unveiled a sweeping plan on Wednesday to \colorbox{mypink}{curb gun violence} that includes a ban on assault weapons and a 10-round limit on high-capacity magazines, as well as measures to improve school safety and mental health. The president's proposal, unveiled at the White House after a month-long review by Vice President Joe Biden and other top officials, \colorbox{mypink}{is expected to face stiff opposition} from gun-control advocates and Republican lawmakers, \colorbox{mypink}{who have vowed to fight it in Congress.}\\\midrule

\textbf{{[}\prop{]}}: President Obama said Wednesday he will propose sweeping gun control measures to \colorbox{mygreen}{address a growing gun violence} crisis in the United States, including a ban on assault-style weapons and requiring background checks for all gun purchases. The proposal was unveiled \colorbox{mygreen}{in response to the latest deadly shootings in America}, the latest in which 20 people were killed by a gunman with an assault-type rifle.\\
% \begin{tabular}[c]{@{}l@{}}\textbf{{[}ModelC{]}}: President Obama said Wednesday he will propose sweeping gun control measures to \colorbox{mygreen}{address a growing gun violence} crisis in the United States, including a ban on assault-style weapons and requiring background checks for all gun purchases. The proposal was unveiled \colorbox{mygreen}{in response to the latest deadly shootings in America}, the latest in which 20 people were killed by a gunman with an assault-type rifle.\end{tabular} \\ 
\midrule \midrule
\textbf{Event 2: Netanyahu plans to deliver a speech to Congress} \\ \midrule
\textbf{{[}\finetune{]}}: Susan Rice, President Obama's national security adviser, said on Tuesday that Israeli Prime Minister Benjamin Netanyahu's planned speech in front of a joint session of Congress will be \colorbox{mypink}{``destructive''} to U.S.-Israel relations. \colorbox{mypink}{``It's destructive to the fabric of the relationship,''} Rice said on PBS's ``Charlie Rose'' show. "We need to keep it that way. The American people want it to be bipartisan." \\ \midrule
\textbf{{[}\prop{]}}: Israel Prime Minister Benjamin Netanyahu will address a joint session of Congress next week on his visit to Washington in what will be his first speech in almost a decade. He will address lawmakers in an address that will come two weeks before he is scheduled to visit the country ahead of his expected election. The event, which \colorbox{mygreen}{will take place at the U.S. Capitol on March 3}, comes amid tensions between President Obama and Republicans in Congress over Iran's nuclear program and tensions between the United States and Israel ahead of next month's elections.\\ 

\bottomrule
\end{tabular}
\caption{Generation examples from \finetune~and \prop~ for two different events. According to our \neutral~metric, \prop~ is \textit{more neutral} than \finetune. We highlight \colorbox{mygreen}{neutral} and \colorbox{mypink}{biased} parts of the generations.}
\label{table:generation_examples}
\end{table*}

\section{Analysis}
\subsection{Decoding Strategy}
In generation tasks, various decoding strategies have been proposed and studied to find the optimal method of decoding for different tasks. For instance, sampling-based decoding techniques are normally adopted to encourage a more diverse response instead of being generic. Thus, we also investigated the impact of decoding strategy on the generation neutrality on the best performing baseline (\prop).

We explored with two focuses: i) whether different decoding strategies affect the neutralization performance, and ii) if yes, what is the best or the worst decoding strategy for the task of neutralization. Four decoding techniques were explored $-$ BEAM Search, Top-k Sampling and Nucleus Sampling (Top-p)\cite{holtzman2019curious}.

The results are shown in Table~\ref{table:decoding_analysis}.  
% shows that decoding strategy does have some impact on the neutralized generation performance. Although there is some variance \textit{within} each decoding techniques depending on the hyper-parameter settings, the variance \textit{among} different decoding techniques are more noticeable. For easy comparison, we report the average scores for each decoding techniques. On average, the top-p decoding technique outperforms other decoding strategies $-$ top-p score is nearly double the greedy score ($15.22 \rightarrow 27.55$). 
For ease of comparison between the decoding techniques, we report the average \neutral~scores as well. We can observe that BEAM and Top-p have rather indifferent average \neutral~scores, although each has its strength - BEAM produces better \rouge~scores, whereas Top-p achieves lower \all~scores ($17\%$). The most notable observation is that the performance drops shown in Top-k decoding results -- they experience marginal drops in \rouge~($10$) while retaining a similar level of bias. It would be an interesting future work to understand the root cause of this phenomenon and devise a decoding technique that can better avoid generating media-bias.

\begin{figure}[t]
    \resizebox{\linewidth}{!}{\includegraphics[width=8cm]{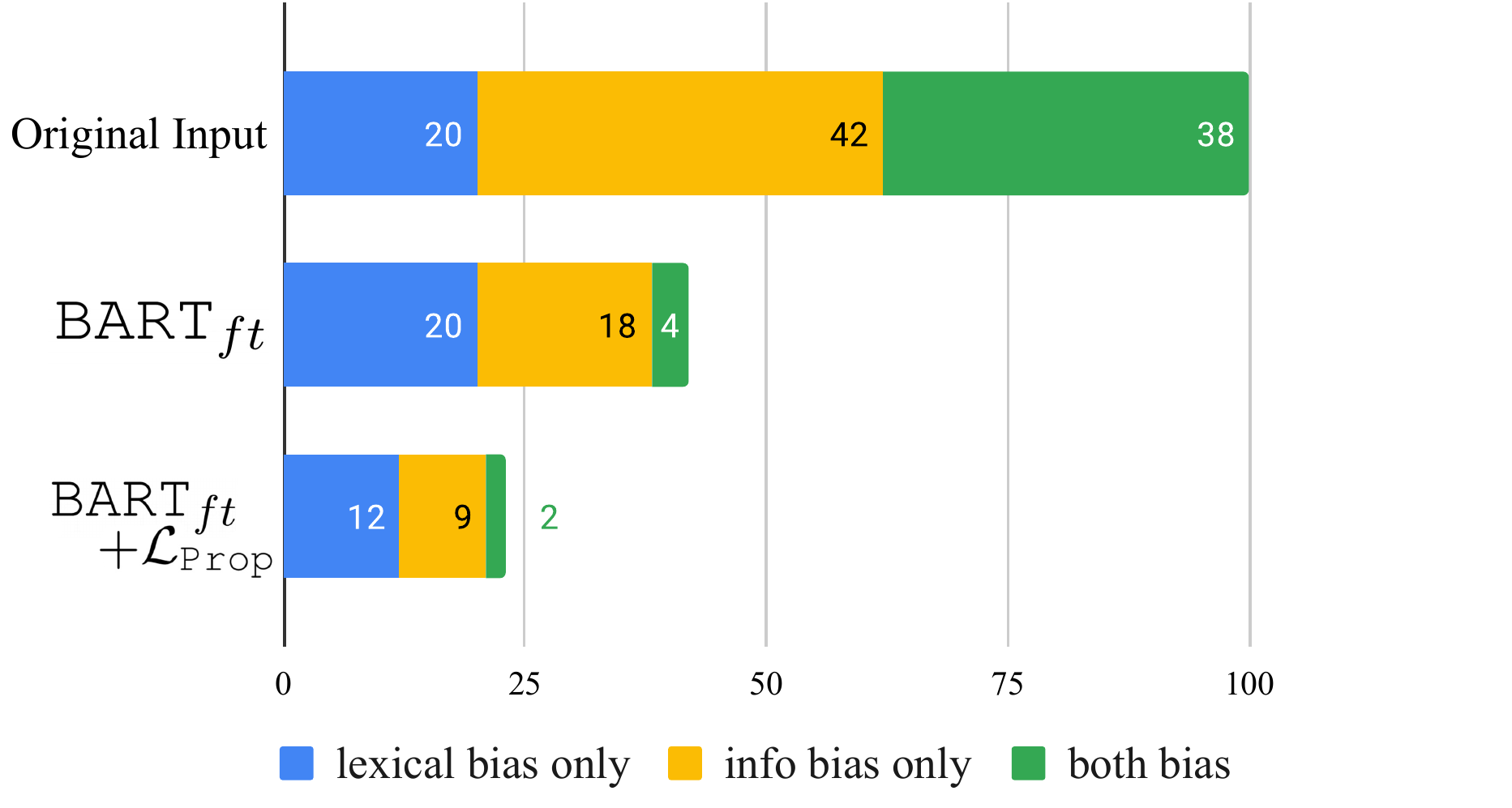}}
    \caption{Illustrating the breakdown of media-bias (lexical vs informational). Original Input refers to biased articles before neutralization. The rest two represent generated articles from \finetune~and \prop~models.}
    % \textbf{FT} and \textbf{FT+disinfo} refer to the generation from \finetune~and \prop~ respectively.}
    \label{fig:bias_breakdown}
\end{figure}

\subsection{Lexical Bias vs. Information Bias}
In Fig~\ref{fig:bias_breakdown}, 
we visualize the breakdown ratio of bias types (lexical bias, informational bias, or both) that exist in different versions of generation.
Through this visualization, we aim to investigate if any one type of bias is harder to eliminate than another. 

To begin with, we illustrate the ratio from ``original input'' that refers to the bias breakdown of the biased input articles ($X^{(l)}+X^{(r)}$). This gives a complete picture of the original breakdown of bias before the neutralization attempt. 
Then, we visualize the bias breakdown of the generation from \finetune~model and \prop~model. 

The most important insight is that majority of the eliminated bias is informational bias. For \prop~, the ``lexical bias only'' ratio is relatively unchanged compared to the big drop in ``info bias only''. Similarly, \finetune's ``lexical bias only'' ratio stays unchanged, clearly indicating the difficulty involved in neutralizing the lexical bias. This is likely because lexical bias normally exists as a very short phrase that is hidden within an otherwise neutral sentence, thus, requiring more sophisticated mitigation.

On the contrary, informational bias is shown to be relatively easy to mitigate. We conjecture this is due to the informational bias often being \textit{additional} piece of information to sway readers mind. There is a high chance for non-overlapping information between politically conflicting articles to be the informational bias -- this serves as a good indicator for the neutralization models.
For instance, as shown in Table \ref{table:basil_example}, \textit{Left} article contains informational bias span ``before President Donald Trump has a chance to interfere,'' negatively targeting Donald Trump, which does not appear in the \textit{Right} article.

\subsection{Generation Examples}
We compare the generation outputs from two models with different 
\neutral~scores to qualitatively check if the difference in scores aligns with the generated articles. In this analysis, we selected two models \prop, relatively more neutral model, and \finetune, which is the less neutral model. 
% $BART\mbox{-}ft$ + $\mathcal{L}_{\texttt{Propaganda}}$, relatively more neutral model, and $BART\mbox{-}ft$, less neutral model, are used. 

The generation examples\footnote{We provide more examples in the appendix.} from Table~\ref{table:generation_examples} illustrates that \prop~indeed generates more neutralized version. 
In the first example about gun control, we can clearly notice a difference in the language used when describing the same information - \prop~writes in a neutral manner (``address'' gun violence''), but \finetune~writes in more sensational style (``curb'' gun violence). Going further, we can also observe the difference in the neutrality from the nature of \textit{additional} information being generated by each model. \prop~provides substantial detail directly related to the event being discussed (i.e., the reason behind the gun control proposal), but \finetune~reports on the negative opposition that ``is expected'' to be faced by those who have ``vowed to fight it in Congress''. 

For the second example related to Netanyahu, \finetune~generates more provoking contents (i.e., the polarized relationship between US and Israel) and lexicons (i.e., ``destructive''), whereas \prop~generation focuses on the factual information related to the actual event (i.e., event will take place at the U.S. Capitol on March 3).

\section{Human Evaluation -- \neutral~metric}
We conduct A/B testing of neutrality between two articles to verify the alignment between our \neutral~metric and human judgment. The A/B pair-comparison method is chosen over the scale-rating evaluations method because it is shown to be more reliable in the literature
\cite{kiritchenko-mohammad-2017-best}. 
We carry out a human evaluation on a data annotation platform, Appen.com. Each annotator is provided two different model generations (i.e., \finetune~ vs. \prop) and is asked to select a less biased version. Our goal is to see if the model with a higher \neutral~score (i.e., \prop) is also perceived to be more neutral by humans as well. 
We obtained three annotations each for 50 random samples of the generations. $74\%$ voted for \prop~ to be \textit{more neutral} with an average sample-wise agreement of $78.14\%$, suggesting that our \neutral~metric aligns with the human judgment. 

In addition, we also asked the annotators to vote for the generation that has higher information overlap with the given neutral article. $60\%$ of the annotators voted for \prop~ to have higher information coverage as well, with the average sample-wise agreement of $75.52\%$.

To ensure the quality of annotation, we did the following. First, we only allow annotators that pass the qualification test step to participate in the A/B testing. Second, we ask for the political orientation to only incorporate the answers from annotators with center political orientation. This is done to avoid as much political bias of annotators as possible. 
% Third, we limit annotator to be U.S. citizens, since they would presumably have a better understanding of the U.S. political context. 
For more detail, refer to the Appendix.

\section{Conclusion}
In this work, we presented a novel generation-based media bias mitigation task. By providing a new dataset, evaluation metric and baselines, this work serves as a solid benchmark for the proposed task. Through a human evaluation, we showed that the proposed metric aligns with the human judgement. Furthermore, our experimental results empirically proved that adding additional disinformation related loss is helpful in neutralization. Lastly, our analysis showed that lexical bias is harder to neutralize than informational bias and that decoding techniques could affect the neutrality of the generations. We hope our work initiates more research on automatic media-bias mitigation to make a positive impact on society.

% \section{Ethical Implication}
% Our proposed work addresses directly a major ethical issue in our society, namely media bias. We propose a new task for computational \textit{mitigation} of media bias. We believe in the potential beneficial impact of our work as, if published, it will lead to more research in mitigating media bias. 
% We also provide a database of media reports from the political left, center, and right. 
% The potential danger of the data set being used to generate more politically polarizing texts is, however, minimal, as the size of the data set is too small for generation purposes. 
% However, the  potential benefit of this work and this data set far out weights any potential misuse. 

\bibliography{acl2020}

\begin{thebibliography}{47}
\expandafter\ifx\csname natexlab\endcsname\relax\def\natexlab#1{#1}\fi

\bibitem[{Bail et~al.(2018)Bail, Argyle, Brown, Bumpus, Chen, Hunzaker, Lee,
  Mann, Merhout, and Volfovsky}]{bail2018exposure}
Christopher~A Bail, Lisa~P Argyle, Taylor~W Brown, John~P Bumpus, Haohan Chen,
  MB~Fallin Hunzaker, Jaemin Lee, Marcus Mann, Friedolin Merhout, and Alexander
  Volfovsky. 2018.
\newblock Exposure to opposing views on social media can increase political
  polarization.
\newblock \emph{Proceedings of the National Academy of Sciences},
  115(37):9216--9221.

\bibitem[{Baum and Groeling(2008)}]{baum2008new}
Matthew~A Baum and Tim Groeling. 2008.
\newblock New media and the polarization of american political discourse.
\newblock \emph{Political Communication}, 25(4):345--365.

\bibitem[{Cao et~al.(2020)Cao, Dong, Wu, and Cheung}]{cao2020factual}
Meng Cao, Yue Dong, Jiapeng Wu, and Jackie Chi~Kit Cheung. 2020.
\newblock Factual error correction for abstractive summarization models.
\newblock \emph{arXiv preprint arXiv:2010.08712}.

\bibitem[{Chiang and Knight(2011)}]{chiang2011media}
Chun-Fang Chiang and Brian Knight. 2011.
\newblock Media bias and influence: Evidence from newspaper endorsements.
\newblock \emph{The Review of economic studies}, 78(3):795--820.

\bibitem[{Da~San~Martino et~al.(2019)Da~San~Martino, Yu, Barr{\'o}n-Cedeno,
  Petrov, and Nakov}]{da2019fine}
Giovanni Da~San~Martino, Seunghak Yu, Alberto Barr{\'o}n-Cedeno, Rostislav
  Petrov, and Preslav Nakov. 2019.
\newblock Fine-grained analysis of propaganda in news article.
\newblock In \emph{Proceedings of the 2019 Conference on Empirical Methods in
  Natural Language Processing and the 9th International Joint Conference on
  Natural Language Processing (EMNLP-IJCNLP)}, pages 5640--5650.

\bibitem[{Dathathri et~al.(2019)Dathathri, Madotto, Lan, Hung, Frank, Molino,
  Yosinski, and Liu}]{dathathri2019plug}
Sumanth Dathathri, Andrea Madotto, Janice Lan, Jane Hung, Eric Frank, Piero
  Molino, Jason Yosinski, and Rosanne Liu. 2019.
\newblock Plug and play language models: a simple approach to controlled text
  generation.
\newblock \emph{arXiv preprint arXiv:1912.02164}.

\bibitem[{De~Vreese(2004)}]{de2004effects}
Claes De~Vreese. 2004.
\newblock The effects of strategic news on political cynicism, issue
  evaluations, and policy support: A two-wave experiment.
\newblock \emph{Mass Communication \& Society}, 7(2):191--214.

\bibitem[{Dong et~al.(2020)Dong, Wang, Gan, Cheng, Cheung, and
  Liu}]{dong2020multi}
Yue Dong, Shuohang Wang, Zhe Gan, Yu~Cheng, Jackie Chi~Kit Cheung, and Jingjing
  Liu. 2020.
\newblock Multi-fact correction in abstractive text summarization.
\newblock \emph{arXiv preprint arXiv:2010.02443}.

\bibitem[{Entman(1993)}]{entman1993framing}
Robert~M Entman. 1993.
\newblock Framing: Toward clarification of a fractured paradigm.

\bibitem[{Entman(2007)}]{entman2007framing}
Robert~M Entman. 2007.
\newblock Framing bias: Media in the distribution of power.
\newblock \emph{Journal of communication}, 57(1):163--173.

\bibitem[{Fan et~al.(2019)Fan, White, Sharma, Su, Choubey, Huang, and
  Wang}]{fan2019plain}
Lisa Fan, Marshall White, Eva Sharma, Ruisi Su, Prafulla~Kumar Choubey, Ruihong
  Huang, and Lu~Wang. 2019.
\newblock In plain sight: Media bias through the lens of factual reporting.
\newblock In \emph{Proceedings of the 2019 Conference on Empirical Methods in
  Natural Language Processing and the 9th International Joint Conference on
  Natural Language Processing (EMNLP-IJCNLP)}, pages 6344--6350.

\bibitem[{Gabriel et~al.(2020)Gabriel, Celikyilmaz, Jha, Choi, and
  Gao}]{gabriel2020go}
Saadia Gabriel, Asli Celikyilmaz, Rahul Jha, Yejin Choi, and Jianfeng Gao.
  2020.
\newblock Go figure! a meta evaluation of factuality in summarization.
\newblock \emph{arXiv preprint arXiv:2010.12834}.

\bibitem[{Gentzkow and Shapiro(2006)}]{gentzkow2006media}
Matthew Gentzkow and Jesse~M Shapiro. 2006.
\newblock Media bias and reputation.
\newblock \emph{Journal of political Economy}, 114(2):280--316.

\bibitem[{Gentzkow and Shapiro(2010)}]{gentzkow2010drives}
Matthew Gentzkow and Jesse~M Shapiro. 2010.
\newblock What drives media slant? evidence from us daily newspapers.
\newblock \emph{Econometrica}, 78(1):35--71.

\bibitem[{Groseclose and Milyo(2005)}]{groseclose2005measure}
Tim Groseclose and Jeffrey Milyo. 2005.
\newblock A measure of media bias.
\newblock \emph{The Quarterly Journal of Economics}, 120(4):1191--1237.

\bibitem[{Hamborg et~al.(2019{\natexlab{a}})Hamborg, Donnay, and
  Gipp}]{hamborg2019automated}
Felix Hamborg, Karsten Donnay, and Bela Gipp. 2019{\natexlab{a}}.
\newblock Automated identification of media bias in news articles: an
  interdisciplinary literature review.
\newblock \emph{International Journal on Digital Libraries}, 20(4):391--415.

\bibitem[{Hamborg et~al.(2017)Hamborg, Meuschke, and Gipp}]{hamborg2017matrix}
Felix Hamborg, Norman Meuschke, and Bela Gipp. 2017.
\newblock Matrix-based news aggregation: exploring different news perspectives.
\newblock In \emph{2017 ACM/IEEE Joint Conference on Digital Libraries (JCDL)},
  pages 1--10. IEEE.

\bibitem[{Hamborg et~al.(2019{\natexlab{b}})Hamborg, Zhukova, and
  Gipp}]{hamborg2019illegal}
Felix Hamborg, Anastasia Zhukova, and Bela Gipp. 2019{\natexlab{b}}.
\newblock Illegal aliens or undocumented immigrants? towards the automated
  identification of bias by word choice and labeling.
\newblock In \emph{International Conference on Information}, pages 179--187.
  Springer.

\bibitem[{Haselmayer and Jenny(2017)}]{haselmayer2017sentiment}
Martin Haselmayer and Marcelo Jenny. 2017.
\newblock Sentiment analysis of political communication: combining a dictionary
  approach with crowdcoding.
\newblock \emph{Quality \& quantity}, 51(6):2623--2646.

\bibitem[{Hermann et~al.(2015)Hermann, Kocisky, Grefenstette, Espeholt, Kay,
  Suleyman, and Blunsom}]{hermann2015teaching}
Karl~Moritz Hermann, Tomas Kocisky, Edward Grefenstette, Lasse Espeholt, Will
  Kay, Mustafa Suleyman, and Phil Blunsom. 2015.
\newblock Teaching machines to read and comprehend.
\newblock \emph{Advances in neural information processing systems},
  28:1693--1701.

\bibitem[{Holtzman et~al.(2019)Holtzman, Buys, Du, Forbes, and
  Choi}]{holtzman2019curious}
Ari Holtzman, Jan Buys, Li~Du, Maxwell Forbes, and Yejin Choi. 2019.
\newblock The curious case of neural text degeneration.
\newblock \emph{arXiv preprint arXiv:1904.09751}.

\bibitem[{Kiritchenko and Mohammad(2017)}]{kiritchenko-mohammad-2017-best}
Svetlana Kiritchenko and Saif Mohammad. 2017.
\newblock \href {https://doi.org/10.18653/v1/P17-2074} {Best-worst scaling more
  reliable than rating scales: A case study on sentiment intensity annotation}.
\newblock In \emph{Proceedings of the 55th Annual Meeting of the Association
  for Computational Linguistics (Volume 2: Short Papers)}, pages 465--470,
  Vancouver, Canada. Association for Computational Linguistics.

\bibitem[{Kry{\'s}ci{\'n}ski et~al.(2019)Kry{\'s}ci{\'n}ski, McCann, Xiong, and
  Socher}]{kryscinski2019evaluating}
Wojciech Kry{\'s}ci{\'n}ski, Bryan McCann, Caiming Xiong, and Richard Socher.
  2019.
\newblock Evaluating the factual consistency of abstractive text summarization.
\newblock \emph{arXiv preprint arXiv:1910.12840}.

\bibitem[{Laban and Hearst(2017)}]{laban2017newslens}
Philippe Laban and Marti~A Hearst. 2017.
\newblock newslens: building and visualizing long-ranging news stories.
\newblock In \emph{Proceedings of the Events and Stories in the News Workshop},
  pages 1--9.

\bibitem[{Lee et~al.(2019)Lee, Liu, and Fung}]{lee2019team}
Nayeon Lee, Zihan Liu, and Pascale Fung. 2019.
\newblock Team yeon-zi at semeval-2019 task 4: Hyperpartisan news detection by
  de-noising weakly-labeled data.
\newblock In \emph{Proceedings of the 13th International Workshop on Semantic
  Evaluation}, pages 1052--1056.

\bibitem[{Lewis et~al.(2019)Lewis, Liu, Goyal, Ghazvininejad, Mohamed, Levy,
  Stoyanov, and Zettlemoyer}]{lewis2019bart}
Mike Lewis, Yinhan Liu, Naman Goyal, Marjan Ghazvininejad, Abdelrahman Mohamed,
  Omer Levy, Ves Stoyanov, and Luke Zettlemoyer. 2019.
\newblock Bart: Denoising sequence-to-sequence pre-training for natural
  language generation, translation, and comprehension.
\newblock \emph{arXiv preprint arXiv:1910.13461}.

\bibitem[{Lin(2004)}]{lin2004rouge}
Chin-Yew Lin. 2004.
\newblock Rouge: A package for automatic evaluation of summaries.
\newblock In \emph{Text summarization branches out}, pages 74--81.

\bibitem[{Lux et~al.(2020)Lux, Sappelli, and Larson}]{lux2020truth}
Klaus-Michael Lux, Maya Sappelli, and Martha Larson. 2020.
\newblock Truth or error? towards systematic analysis of factual errors in
  abstractive summaries.
\newblock In \emph{Proceedings of the First Workshop on Evaluation and
  Comparison of NLP Systems}, pages 1--10.

\bibitem[{Ma et~al.(2020)Ma, Sap, Rashkin, and Choi}]{ma2020powertransformer}
Xinyao Ma, Maarten Sap, Hannah Rashkin, and Yejin Choi. 2020.
\newblock Powertransformer: Unsupervised controllable revision for biased
  language correction.
\newblock In \emph{Proceedings of the 2020 Conference on Empirical Methods in
  Natural Language Processing (EMNLP)}, pages 7426--7441.

\bibitem[{McCombs and Reynolds(2009)}]{mccombs2009news}
Maxwell McCombs and Amy Reynolds. 2009.
\newblock How the news shapes our civic agenda.
\newblock In \emph{Media effects}, pages 17--32. Routledge.

\bibitem[{Miller and Riechert(2001)}]{miller2001spiral}
M~Mark Miller and Bonnie~Parnell Riechert. 2001.
\newblock The spiral of opportunity and frame resonance: Mapping the issue
  cycle in news and public discourse.
\newblock \emph{Framing public life: Perspectives on media and our
  understanding of the social world}, pages 107--121.

\bibitem[{Narayan et~al.(2018)Narayan, Cohen, and Lapata}]{narayan2018don}
Shashi Narayan, Shay~B Cohen, and Mirella Lapata. 2018.
\newblock Don’t give me the details, just the summary! topic-aware
  convolutional neural networks for extreme summarization.
\newblock In \emph{Proceedings of the 2018 Conference on Empirical Methods in
  Natural Language Processing}, pages 1797--1807.

\bibitem[{Park et~al.(2009)Park, Kang, Chung, and Song}]{park2009newscube}
Souneil Park, Seungwoo Kang, Sangyoung Chung, and Junehwa Song. 2009.
\newblock Newscube: delivering multiple aspects of news to mitigate media bias.
\newblock In \emph{Proceedings of the SIGCHI conference on human factors in
  computing systems}, pages 443--452.

\bibitem[{Park et~al.(2011)Park, Lee, and Song}]{park2011contrasting}
Souneil Park, Kyung-Soon Lee, and Junehwa Song. 2011.
\newblock Contrasting opposing views of news articles on contentious issues.
\newblock In \emph{Proceedings of the 49th annual meeting of the association
  for computational linguistics: Human language technologies}, pages 340--349.

\bibitem[{Perse and Lambe(2016)}]{perse2016media}
Elizabeth~M Perse and Jennifer Lambe. 2016.
\newblock \emph{Media effects and society}.
\newblock Routledge.

\bibitem[{Potthast et~al.(2017)Potthast, Kiesel, Reinartz, Bevendorff, and
  Stein}]{potthast2017stylometric}
Martin Potthast, Johannes Kiesel, Kevin Reinartz, Janek Bevendorff, and Benno
  Stein. 2017.
\newblock A stylometric inquiry into hyperpartisan and fake news.
\newblock \emph{arXiv preprint arXiv:1702.05638}.

\bibitem[{Pryzant et~al.(2020)Pryzant, Martinez, Dass, Kurohashi, Jurafsky, and
  Yang}]{pryzant2020automatically}
Reid Pryzant, Richard~Diehl Martinez, Nathan Dass, Sadao Kurohashi, Dan
  Jurafsky, and Diyi Yang. 2020.
\newblock Automatically neutralizing subjective bias in text.
\newblock In \emph{Proceedings of the AAAI Conference on Artificial
  Intelligence}, volume~34, pages 480--489.

\bibitem[{Puglisi(2011)}]{puglisi2011being}
Riccardo Puglisi. 2011.
\newblock Being the new york times: the political behaviour of a newspaper.
\newblock \emph{The BE journal of economic analysis \& policy}, 11(1).

\bibitem[{Raffel et~al.(2019)Raffel, Shazeer, Roberts, Lee, Narang, Matena,
  Zhou, Li, and Liu}]{raffel2019exploring}
Colin Raffel, Noam Shazeer, Adam Roberts, Katherine Lee, Sharan Narang, Michael
  Matena, Yanqi Zhou, Wei Li, and Peter~J Liu. 2019.
\newblock Exploring the limits of transfer learning with a unified text-to-text
  transformer.
\newblock \emph{arXiv preprint arXiv:1910.10683}.

\bibitem[{Recasens et~al.(2013)Recasens, Danescu-Niculescu-Mizil, and
  Jurafsky}]{recasens2013linguistic}
Marta Recasens, Cristian Danescu-Niculescu-Mizil, and Dan Jurafsky. 2013.
\newblock Linguistic models for analyzing and detecting biased language.
\newblock In \emph{Proceedings of the 51st Annual Meeting of the Association
  for Computational Linguistics (Volume 1: Long Papers)}, pages 1650--1659.

\bibitem[{Sides(2018)}]{sides2018media}
All Sides. 2018.
\newblock Media bias ratings.
\newblock \emph{Allsides.com}.

\bibitem[{Vaswani et~al.(2017)Vaswani, Shazeer, Parmar, Uszkoreit, Jones,
  Gomez, Kaiser, and Polosukhin}]{vaswani2017attention}
Ashish Vaswani, Noam Shazeer, Niki Parmar, Jakob Uszkoreit, Llion Jones,
  Aidan~N Gomez, {\L}ukasz Kaiser, and Illia Polosukhin. 2017.
\newblock Attention is all you need.
\newblock In \emph{Advances in neural information processing systems}, pages
  5998--6008.

\bibitem[{Wolf et~al.(2020)Wolf, Debut, Sanh, Chaumond, Delangue, Moi, Cistac,
  Rault, Louf, Funtowicz, Davison, Shleifer, von Platen, Ma, Jernite, Plu, Xu,
  Scao, Gugger, Drame, Lhoest, and Rush}]{wolf-etal-2020-transformers}
Thomas Wolf, Lysandre Debut, Victor Sanh, Julien Chaumond, Clement Delangue,
  Anthony Moi, Pierric Cistac, Tim Rault, Rémi Louf, Morgan Funtowicz, Joe
  Davison, Sam Shleifer, Patrick von Platen, Clara Ma, Yacine Jernite, Julien
  Plu, Canwen Xu, Teven~Le Scao, Sylvain Gugger, Mariama Drame, Quentin Lhoest,
  and Alexander~M. Rush. 2020.
\newblock \href {https://www.aclweb.org/anthology/2020.emnlp-demos.6}
  {Transformers: State-of-the-art natural language processing}.
\newblock In \emph{Proceedings of the 2020 Conference on Empirical Methods in
  Natural Language Processing: System Demonstrations}, pages 38--45, Online.
  Association for Computational Linguistics.

\bibitem[{Yano et~al.(2010)Yano, Resnik, and Smith}]{yano2010shedding}
Tae Yano, Philip Resnik, and Noah~A Smith. 2010.
\newblock Shedding (a thousand points of) light on biased language.
\newblock In \emph{Proceedings of the NAACL HLT 2010 Workshop on Creating
  Speech and Language Data with Amazon’s Mechanical Turk}, pages 152--158.

\bibitem[{Zhang et~al.(2019)Zhang, Da~San~Martino, Barr{\'o}n-Cedeno, Romeo,
  An, Kwak, Staykovski, Jaradat, Karadzhov, Baly et~al.}]{zhang2019tanbih}
Yifan Zhang, Giovanni Da~San~Martino, Alberto Barr{\'o}n-Cedeno, Salvatore
  Romeo, Jisun An, Haewoon Kwak, Todor Staykovski, Israa Jaradat, Georgi
  Karadzhov, Ramy Baly, et~al. 2019.
\newblock Tanbih: Get to know what you are reading.
\newblock \emph{EMNLP-IJCNLP 2019}, page 223.

\bibitem[{Zhao et~al.(2020)Zhao, Cohen, and Webber}]{zhao2020reducing}
Zheng Zhao, Shay~B Cohen, and Bonnie Webber. 2020.
\newblock Reducing quantity hallucinations in abstractive summarization.
\newblock \emph{arXiv preprint arXiv:2009.13312}.

\bibitem[{Zhou et~al.(2020)Zhou, Gu, Diab, Guzman, Zettlemoyer, and
  Ghazvininejad}]{zhou2020detecting}
Chunting Zhou, Jiatao Gu, Mona Diab, Paco Guzman, Luke Zettlemoyer, and Marjan
  Ghazvininejad. 2020.
\newblock Detecting hallucinated content in conditional neural sequence
  generation.
\newblock \emph{arXiv preprint arXiv:2011.02593}.

\end{thebibliography}
\bibliographystyle{acl_natbib}

\end{document}

\maketitle
\appendix

\section{Ethical Implication}
Our proposed work addresses directly a major ethical issue in our society, namely media bias. We propose a new task for computational \textit{mitigation} of media bias. We believe in the potential beneficial impact of our work as, if published, it will lead to more research in mitigating media bias. 

\section{\rougetwo~and \rougeL}
\begin{table}[h]
\resizebox{\linewidth}{!}{
\centering
\begin{tabular}{llcc}
\toprule
 & Baseline & \rougetwo & \rougeL  \\ \midrule \midrule
\multirow{2}{*}{A. Zero-shot} 
& \bartCNN & 5.29 & 10.31 \\
& \bartXSum & 2.66 & 5.4 \\ 
\midrule
% \multirow{2}{*}{B. Fine-tuning Seq2Seq} 
B. Fine-tuning & \finetune~ & 13.78 & 20.20 \\
Seq2Seq & \tfive~ & 6.97 & 11.81 \\ 
 \midrule
% \multirow{2}{*}{C. \finetune + Disinfo Loss} 
C. \finetune& \fakenews & 16.27 & 23.52 \\
 + Disinfo Loss& \prop & 15.25 & 22.59  \\
 \bottomrule
\end{tabular}}
\caption{\rougetwo~ and \rougeL~ recall scores for baseline generations using beam-4 decoding.}
\label{table:baseline_rougemore}
\end{table}

\section{Human Evaluation Detail}
\subsection{Human Evaluation Interface and Task}
The human evaluation was to check alignment of our metric and actual generated articles. Before annotators are given the set of articles, they need to read instruction to have basic understanding of media bias. Please refer to Figure \ref{fig:annotation_inst}.

Annotators are asked to choose between given articles that are anonymously names as ``Article 1'' and ``Article 2''. The questions that annotators were asked to answer are following:
\begin{enumerate}
   \item Which article is more \textit{biased}?
   \item Which article is more \textit{neutral}?
   \item Which article contains \textit{more} information that overlapping with the given neutral article?
   \item Which article contains \textit{less} information that overlapping with the given neutral article?
   \item Check if you are a citizen of U.S.
 \end{enumerate}

\subsection{Human Evaluation Quality Control}
To ensure the quality of human evaluation, we only selected the qualified annotators who are experienced, possessing higher accuracy rate in the annotation platform (Appen.com).

We also added a separate set of qualification tasks so we could ensure the human annotator understand the task. We designed the quiz set by taking human-written neutral summary from Allsides.com to be a neutral article. Then, we prepared two manually selected articles (i) Neutral-Modified article, which is expected to be more neutral (ii) Biased article, which is expected to be selected as more biased.
For Neutral-Modified article, we changed the sequence of sentence, but keeping the information and style of writing unchanged. For the biased sample, we took opinion pieces\footnote{\textit{Opinion piece} is an article, usually published in a newspaper, that mainly reflects the author's opinion about a subject.} written about the event. An example qualification quiz set is available in Table \ref{table:quality_control}.

\begin{table*}[t]
    \begin{minipage}{.5\linewidth}
        \begingroup
             \includegraphics[width=\linewidth]{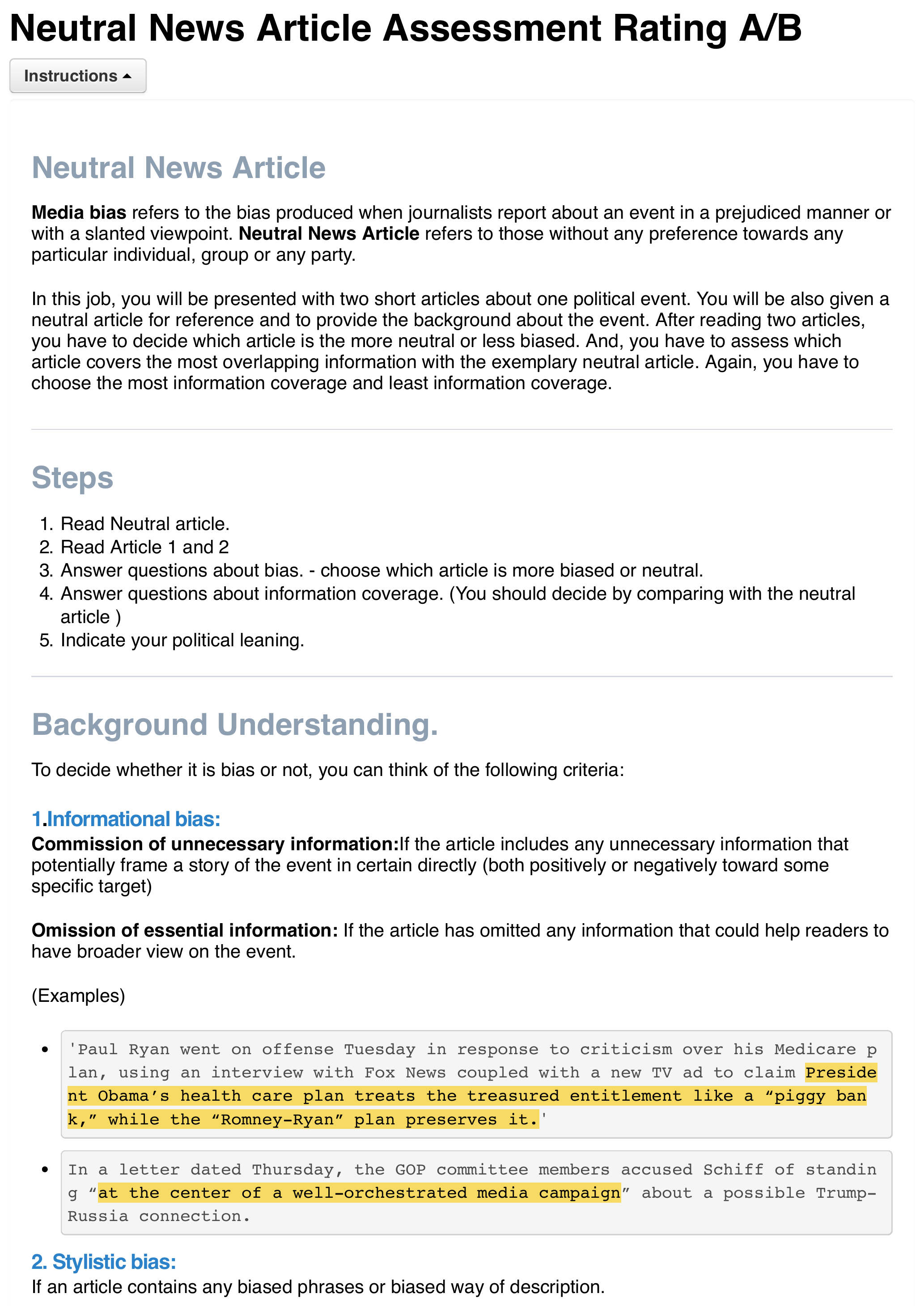}
        \vspace{-20pt}
        % \captionof{figure}{Plot of experiment results. The closer to the top-right, the better.}
        \label{fig:human_Eval}
        \endgroup
    \end{minipage}%
    \hspace{8pt}
    \begin{minipage}{.5\linewidth}
        \centering
        \begingroup
        \includegraphics[width=\linewidth]{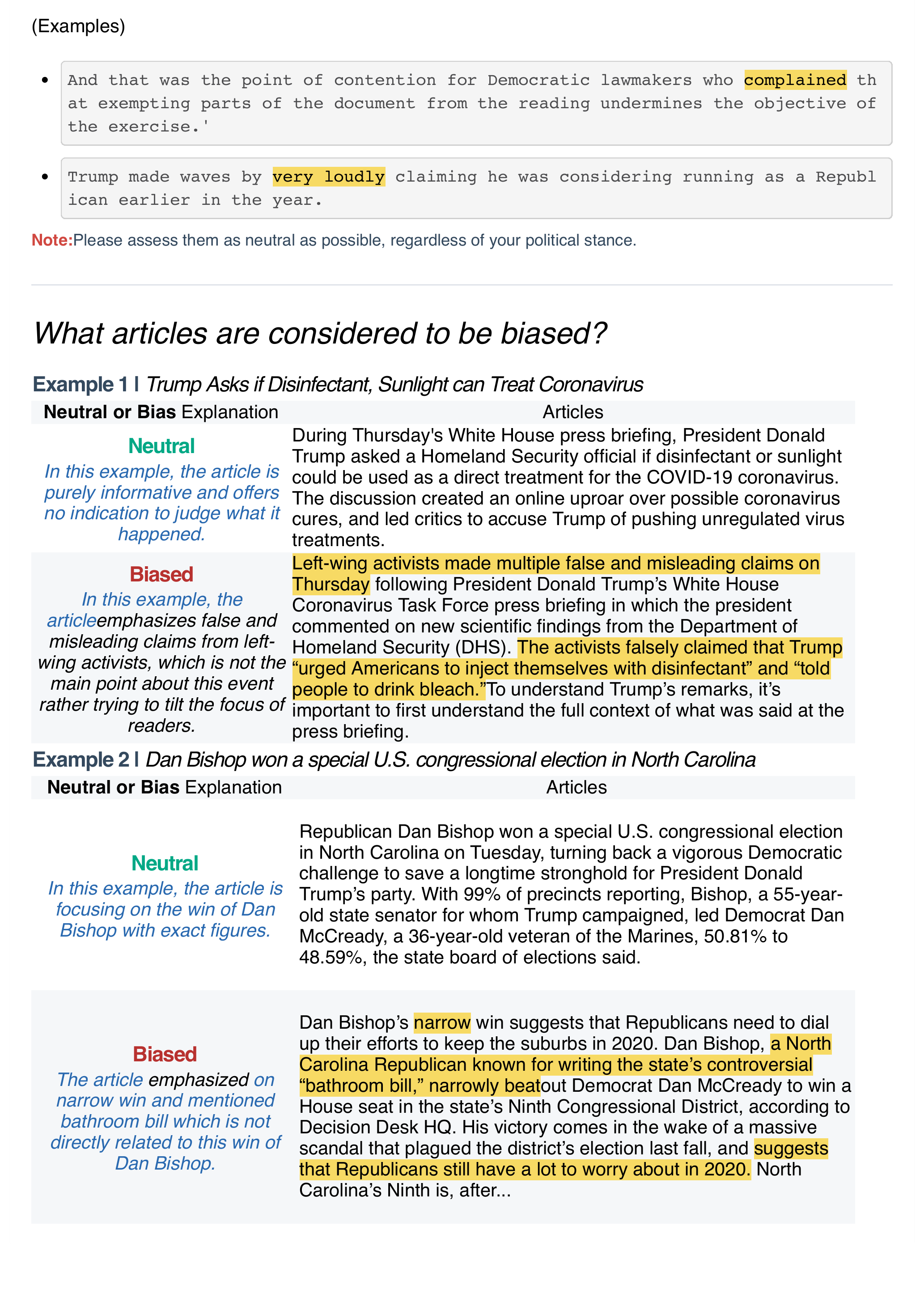}
        \vspace{-20pt}
        % \captionof{figure}{Plot of experiment results. The closer to the top-right, the better.}
        \label{fig:result_illustration}
        \endgroup
    \end{minipage}%
    \captionof{figure}{Screenshot of instruction for human evaluation annotation. After this instruction, annotators are given a set of articles and need to answer questions.}
    \label{fig:annotation_inst}
\end{table*}

\begin{table*}[t]
\resizebox{\linewidth}{!}{

    \centering
    \begin{tabular}{c|p{15cm}}
\toprule
 \multicolumn{2}{c}{\texttt{Qualification Task Set}}\\\midrule
        \multicolumn{2}{l}{\textbf{Event: White House Legal Team Decries Alleged Election Fraud, Projects Trump as Winner}}\\\midrule
        \textbf{Neutral} & President Donald Trump's lawyer Rudy Giuliani and former federal prosecutor Sidney Powell led a press conference Thursday, claiming to have proof of a mass conspiracy to elect Joe Biden and manipulate the 2020 presidential election results against Trump. Trump's legal team described him as the election's winner on Thursday; Joe Biden is widely projected to be elected the next president, and there's currently no public evidence of the decisive voter fraud claimed by Trump, Giuliani and others.\\\midrule\midrule
        % \multicolumn{2}{l}{Article 1}\\\midrule
        \multirow{2}{*}{\textbf{Neutral Modified}} &\textbf{Article 1: }\\
        & Rudy Giuliani, President Donald Trump's lawyer, and Sidney Powell, former federal prosecutor, led a press conference, claiming to have proof of a mass conspiracy to elect Joe Biden and manipulate the 2020 presidential election results against Trump. Trump's legal team described him as the election's winner on Thursday. Currently, there is no public evidence of the decisive voter fraud claimed by Trump, Giuliani and others while Joe Biden is widely projected to be elected the next president.\\\midrule
        % \multicolumn{2}{l}{Article 2}\\\midrule
        \multirow{2}{*}{\textbf{Biased}} &\textbf{Article 2: }\\
        &  Rudy Giuliani sweated so much during a press conference Thursday that streaks of what appeared to be dark hair dye streamed down his cheeks, making for a bizarre juxtaposition as the ex-mayor insisted without evidence that Democrats stole the election from President Trump. Giuliani, who’s spearheading the Trump campaign’s bumbling legal battle to overturn Joe Biden’s election, did not appear to notice the black liquid dripping down from his sideburns, though he kept dabbing pearls of sweat from his face with a napkin.\\
        \bottomrule
\end{tabular}
}
\caption{Qualification task article sets for human evaluation. Only the annotators who correctly distinguish the biased article (opinion piece) from neutral modified one could pass it.}
\label{table:quality_control}
\end{table*}

\section{More examples of generations}
We provide more example generations from \finetune~and \prop~. Please refer to Table \ref{tab:generation examples} in page 3 and 4. Due to the limited space, the examples are short version of generations.

\begin{table*}[t]
    \centering
    \begin{tabular}{p{15cm}}
    \toprule
    \texttt{Neutralized Example 1}\\\midrule
        \textbf{Event: Nebraska and Oklahoma fight Colorado's recreational marijuana policy}\\\midrule
        \textbf{[\prop]} Colorado's recreational marijuana law is under attack by two neighboring states, which on Thursday sued to stop the state from enforcing its state law. In a federal lawsuit filed in the U.S. Supreme Court, attorneys general from Nebraska and Oklahoma accuse the state of acting illegally by allowing marijuana sales within its borders under its so-called Amendment 64, a measure voters overwhelmingly approved in 2012 that legalizes the possession and consumption of cannabis for adults. The two states filed a lawsuit in federal court in Nebraska asking the court to stop Colorado from enforcing the law.\\\midrule
        \textbf{[\finetune]} Nebraska and Oklahoma on Thursday filed a federal lawsuit challenging Colorado's legalization of recreational marijuana, saying the state's law violates federal law. The two states, which voted overwhelmingly in 2012 to legalize the drug for adults over 21, say the measure violates the supremacy clause of the U.S. Constitution, which bars the federal government from interfering with state marijuana laws. "Federal law undisputedly prohibits the production and sale of marijuana," Nebraska Attorney General Jon Bruning said at a news conference in Lincoln.\\\bottomrule
    \end{tabular}
    % \caption{Caption}
    % \label{tab:my_label}
     \centering
    \begin{tabular}{p{15cm}}
    \\
    \toprule
        \texttt{Neutralized Example 2}\\\midrule

        \textbf{Event: AARP drops its opposition to Social Security cuts}\\\midrule
        \textbf{[\prop]} The head of the American Association of Retired Americans told CNN on Friday he had no plans to change its decades-old stance on cuts to Social Security benefits, a move that would signal a shift in the organization's position on the government's entitlement program. David Certner said the group's position remains unchanged on cuts, despite a Wall Street Journal report on Thursday that it could accept cuts in benefits under a potential privatization proposal that would shrink benefits and increase taxes on wealthier Americans.\\\midrule
        \textbf{[\finetune]} WASHINGTON (AP) - The head of the AARP, the nation's largest advocacy group for senior citizens, on Friday disputed a report that said the group was softening its opposition to cutting Social Security benefits. AARP CEO Barry Rand issued a statement saying the Wall Street Journal story was "misleading" and that the group remains opposed to cutting benefits for the sake of deficit reduction and to any privatization plan.\\\bottomrule
    \end{tabular}
    \begin{tabular}{p{15cm}}
    \\
    \toprule
            \texttt{Neutralized Example 3}\\\midrule
        \textbf{Event: White House Plays Down Data Program}\\\midrule
        \textbf{[\prop]} The president on Friday defended his administration's role in intelligence programs as part of the "war on terrorism" after the Guardian and The Washington Post reported on a secret phone records collection program that spied on Internet companies. "I welcome this debate and I think it's healthy for the American people to have a full and full discussion," President Barack Obama said. "The programs that we have in place are essential for keeping us safe from terrorism and keep us in the know about the threat of radicalization and foreign interference”\\\midrule
        \textbf{[\finetune]} President Obama on Friday defended his administration's surveillance programs, saying they are essential to fighting terrorism and that Congress has been fully briefed on them. "The programs are secret in the sense that they are classified. They are not secret, in that every member of Congress have been briefed," he said during a speech in San Jose, Calif. "These are programs that have been authored by large bipartisan majorities repeatedly since 2006."\\\bottomrule
    \end{tabular}

\end{table*}

\begin{table*}[t]
    \centering
    \begin{tabular}{p{15cm}}
    \toprule
    \texttt{Neutralized Example 4}\\\midrule

        \textbf{Event: Walmart sex discrimination case goes to supreme court}\\\midrule
        \textbf{[\prop]} A group of women alleging gender discrimination at Wal-Mart Stores Inc. is seeking a nationwide class action lawsuit over pay and promotions that could lead to billions in damages from the world's largest retailer. The Supreme Court heard arguments on Tuesday about whether the company has violated the rights of workers who claim they were underpaid or mistreated at work. The nine justices heard arguments Tuesday before a panel of three liberal justices about whether Wal-MEX Inc. must answer claims that it pays its workers less than men and sets wage and job conditions that favor men.\\\midrule
        \textbf{[\finetune]} The Supreme Court on Tuesday appeared ready to dismiss a massive class action lawsuit against Wal-Mart alleging that the world's largest retailer discriminates against women by favoring men in pay and promotions. The justices heard arguments in a case that could be the largest class action of its kind in U.S. history and could have far-reaching implications for other women's rights cases. Justice Anthony Kennedy, often a key vote on the high court, said the women's argument points in apparently conflicting directions.\\\bottomrule
    \end{tabular}
    \caption{Neutralized article examples for four different events from \finetune~and~\prop~models.}
    \label{tab:generation examples}
\end{table*}